\documentclass[10pt,onecolumn]{article}
\pdfoutput = 1
\usepackage{graphicx}
\usepackage{subfigure}
\usepackage{tabularx}

\begin{document}
\title{G-Lets: Signal Processing using Transformation Groups}
\author{B.Rajathilagam, Murali Rangarajan,  K.P.Soman\\Amrita Vishwa Vidyapeetham, Coimbatore,\\India}

\maketitle

\begin{abstract}

We present an algorithm using transformation groups and their irreducible representations to generate an orthogonal basis for a signal in the vector  space of the signal. It is shown that multiresolution analysis can be done with amplitudes using a transformation group. G-lets is thus not a single transform, but a group of linear transformations related by group theory. The algorithm also specifies that a multiresolution and multiscale analysis for each resolution is possible in terms of frequencies. Separation of low and high frequency components of each amplitude resolution is facilitated by G-lets. Using conjugacy classes of the transformation group, more than one set of basis may be generated, giving a different perspective of the signal through each basis. Applications for this algorithm include edge detection, feature extraction, denoising, face recognition, compression, and more. We analyze this algorithm using dihedral groups as an example. We demonstrate the results with an ECG signal and the standard `Lena' image.

\end{abstract}

\subsection*{keywords}
 representations, group theory, signal processing, orthogonal basis, signal decomposition, wavelets, multiresolution analysis, multidimensional signal processing

\section{Introduction}
A set of transformations, using group theory, provides an interesting framework with unique properties that may be exploited to analyze a signal when mapped to the vector space of the signal. The choice of a group, and the resultant set of transformations, provides a natural basis for a signal. The relationship between the transformations allows for a multiresolution analysis of the signal. The basis set, in group theory~\cite{Ref:Riley}~\cite{Ref:Schur}~\cite{Ref:Duzhin}~\cite{Ref:Hamermesh}, is not a set of points in Euclidean space, but the set of transformations in the space. This way, a signal can be analyzed through a set of transformations in the vector space that it occupies. Linear transformations of a vector space are many, including rotation, translation, reflection, and more. Every transformation group follows certain common rules and has some unique properties, which can be used to portray the structural symmetries of a signal. In this analysis, we represent the transformations as sparse matrix representations. Our basis is called G-lets for two reasons: one because we use group theory; the other because it is a generalized form of both Fourier and Wavelet analysis. A basic introduction to G-lets is found in the reference~\cite{Ref:thilagam}.

The local information of a signal is spread across the basis set in a Fourier analysis of a signal~\cite{Ref:Hamermesh}. On the other hand, using multiresolution analysis, wavelets are able to collect low frequency information at large amplitudes and high frequency information at small amplitudes. Use of more dilated wavelet basis functions produces a blurring and ringing noise at the edge regions in images~\cite{Ref:Mallatwavelet}. There is still scope for exploration in the extraction of localized information in a signal. In the present work, using group theory, we have created a set of sparse matrix representations of a signal using different transformations. Each basis generated out of this set exhibits different amplitudes and frequencies, corresponding to a particular resolution of the signal. This happens gradually through each element of the basis, clearly identifying the areas of abrupt variations. Thus, there is no necessity for the choice of an approximate signal like a mother wavelet to capture the signal's frequencies.

\section{Background}

Capturing the local information in various types of signals has been explored by many transforms. Non-stationary signals are poorly represented by a Fourier transform. It also has a problem with discontinuous signals due to Gibbs effect. The short-time Fourier transforms and other generalized Fourier transforms like Fokas transform~\cite{Ref:Fokas} are not able to manage efficiently both the time and frequency elements for local features in a signal. Wavelets~\cite{Ref:Mallatwavelet} ~\cite{Ref:Daubechies-1988} ~\cite{Ref:Daubechies} grab both the time and frequency behavior of a signal in steps of multiple scales. Some intricate features of a signal cannot be understood by self similarity and long range dependence of wavelets. As the dilation of wavelets increases blurring of the edges in an image is unavoidable. In Geometric wavelets~\cite{Ref:Chen} a \textit{d}-dimensional plane fit to the data with axes in the directions of maximum variance is used similar to a scaling function in wavelets. Projection of data on this plane gives coarse projections. A principal component analysis is performed on the finer scale data and using the differences between these two, decomposition of the signal is obtained. Thus Geometric wavelets provide a multiscale dictionary or feature set of the data that efficiently captures coarse-to-fine structure in the data. They are used for compressing databases and analyzing online data. Ridgelet transform~\cite{Ref:Candesphd}~\cite{Ref:Candes} focuses on detecting lines in an image using Radon transform~\cite{Ref:Matus}, and basically it is a wavelet transform. The curves are specially recognized in a signal by curvelets~\cite{Ref:Donoho} which is a combination of wavelets and ridgelets. These two transforms are specialized forms of wavelets. There are also other specialized forms of wavelets like the contourlets~\cite{Ref:Do}, Wedgelets~\cite{Ref:Donohowedgelets}, and Grouplets~\cite{Ref:Mallatgrouplet}. Representations of transformations like rotations and reflections have been used by Lenz for edge detection in images in three dimensions using correlation as a parameter. In his work, Lenz~\cite{Ref:Lenz-1990} ~\cite{Ref:Lenz-2009} uses octahedral transformation group in the 3-D image environment, and the representations have been used to describe this particular environment. In another paper, Vale~\cite{Ref:Vale} shows how an orthonormal set of polynomials can be generated by the orbit of a vector using representations. Sparse representation~\cite{Ref:Starck} are discussed in terms of overcomplete dictionaries~\cite{Ref:Aharon} where the discussion is on generating a sparse representation basis from a dictionary of different basis sets of the signal. In this work, we focus on providing a generalized framework using transformation groups for capturing the structural symmetries of a signal through its amplitudes and frequencies. We find that a natural way of multiresolution analysis emerges.

\section{Transformations for G-lets basis}

We choose the dihedral Groups $D$~\cite{Ref:Riley}~\cite{Ref:Lenz-1990}~\cite{Ref:Viana}~\cite{Ref:Dresselhaus} for demonstration, which consist of rotations and reflections, and show its properties to construct G-lets. We consider only discrete signals. An illustration of using group theory for generating G-lets with simple examples of $3$, $6$ and $9$ tuples are presented in the reference~\cite{Ref:thilagam}. Using the theory of representations we can associate one matrix representation to each of the transformations in the group. If the representations are unitary, they can be divided into sub-representations called the irreducible representations. These irreducible representations form the full matrix of the representation. The irreducible representations also form an orthogonal basis of the vector space whose dimension is that of the group.
An interesting feature of transformation groups~\cite{Ref:Riley}~\cite{Ref:Hamermesh} is that they contain conjugacy classes. A conjugacy class contains transformations in a group that are related only if there exists another transformation P such that,
\begin{equation}
P M P^{-1} = N
\label{Eq:conjugacy}
\end{equation}

where M, N, and P are elements of the group. Here M and N are said to be conjugate elements of the group. Thus, the conjugacy class contains all transformations which are conjugate to one another. The irreducible representations of a transformation are unique only outside a conjugacy class. Consequently, the number of irreducible representations is equal to the number of conjugacy classes. The dimension of an irreducible representation is the same as the number of transformations in the corresponding conjugacy class. Conjugate transformations expose different structural symmetries or viewpoints of the signal. So there are as many basis sets as there are conjugate members in each class.

To construct a G-let basis, we choose the dihedral group with the same dimension as the dimension of the discrete signal `n'. Then the group has n rotations$(R)$ and n reflections$(S)$. They form a group with angle of $\theta = 360^\circ/n$ and the following relationship:

\begin{equation}
G = \{ R, R^2, R^3, R^4,{\cdots}R^n, S, SR, SR^2, SR^3, {\cdots}SR^{n-1}\}
\label{Eq:dihedralgroup}
\end{equation}

There are three relations between the generating elements R and S of the above group. They are
\begin{equation}
S^2 = id; \; {(S {\times} R)}^2 =id; \; R^n = id;
\label{Eq:groupconditions}
\end{equation}

where id = identity transformation. Each transformation is connected with a single representation matrix. The operation is defined to be multiplication for a matrix representation. Rotations make one G-let basis and reflections create another G-let basis. To calculate the irreducible representations, we first find the conjugacy classes. It is known that if n is odd, then the number of conjugacy classes is $(n+3)/2$, whereas if n is even, then it is $(n+6)/2$. The identity transformation forms a conjugacy class by itself. The other conjugate members are formed by $S{\times}R^k{\times}S = R^{-k}$. Hence the conjugacy classes are ${ R^n }, {R, R^2 }, {SR, SR^2, SR^3, SR^{n-1} }$ for an odd dimension `n' of the signal. For the odd dimension, all the reflections make one conjugacy class. The classes for an even dimension are ${R}, { R^n }, {R, R^2 }, {SR, SR^{n/2} } \textrm{and  } { SR^{n/2+1},  SR^{n-1}}$. The reflections are grouped into two conjugacy classes. As a result the irreducible representations are just one for each conjugacy class. We thus have two basis sets in both cases of odd and even dimensions of a signal, by choosing only one member from each conjugacy class for one set. Every rotation (or reflection) basis vector is a sparse diagonal representation matrix.

The identity representation $(id)$ does not produce any change in the signal when projected. So we only use these one dimensional irreducible representations to match the dimension of the representation matrix with that of the signal. For example, the representation matrix for rotation is given by the $i^{th}$ irreducible representation $(r_i)$ as
\begin{equation}
r_i = \left(
\begin{array}{clrr}%
 \cos i\times\theta&-\sin i\times\theta  \\ \sin i\times\theta&\cos i\times\theta
\end{array}
\right)
\label{Eq:rotation}
\end{equation}

and the reflection matrix $(s_i)$ is any random reflection matrix. The sparse diagonal representation matrix is then built by placing the corresponding irreducible representations diagonally. For a dihedral group, the dimension is two for $r_i$ and $s_i$ and the full representation matrix is given by
\begin{equation}
\left(
\begin{array}{clrr}%
id \qquad \qquad \qquad \qquad 0\\ R = r_i \\ \qquad \qquad \qquad R = r_i \\ \qquad 0 \qquad \qquad \qquad \qquad \ddots
\end{array}
\right)
\label{Eq:rotation_matrix}
\end{equation}

\section{Construction of G-Lets}

The signal is now projected onto the G-let basis. The basis gradually filters from low pass to high pass, the amplitudes of the signal, presenting a different amplitude resolution for each G-let. For an $i^{th}$ G-let, let $R^i$ be the corresponding matrix representation for a given signal \textit{sig}. Then,
\begin{equation}
G_i(sig) = R_i
\end{equation}

In order to generate frequencies, each G-let is split into a low $Low_i$ and high $High_i$ frequency component. A linear sum of these two components gives the related G-let coefficients as shown below.
\begin{equation}		  	
		G_i(sig) = Low_i(sig) \oplus High_i(sig)
\end{equation}

The algorithm for the generation of G-lets, G-let matrices and the reconstruction is shown below.

\subsection{\textit{G-let\_Decomposition} (sig, D)}

\begin{itemize}
        \item \em Calculate the dimension `n' of the signal `sig'
        \item Choose a transformation group, for example the Dihedral group D
        \item Generate the basis matrices for G-let G(sig) from $1$ to $2n$ using \\`Generate\_G-let\_Matrices' function given below. \\The $i^{th}$ G-let is taken as $G_i(sig) = R_i$
        \item Project the signal onto the G-let matrices to oscillate the signal
        \item Each G-let is a different resolution in terms of amplitude produced by the corresponding oscillations
        \item For each amplitude resolution, obtain low and high frequencies by removing the oscillatory portion of the signal
        \item The signal is now a linear sum of the G-let coefficients \\
                \\sig  = $(G_1(sig) \oplus G_2(sig) \oplus G_3(sig) \cdots \oplus Gn-1(sig))\times(-In)$ \\
                sig =  $(G_{n+1}(sig) \oplus G_{n+2}(sig) \oplus G_{n+3}(sig) \cdots \oplus G_{n-1}(sig))\times(-In)$ \\
                \\$\qquad\qquad\qquad$ These are the first and second basis sets corresponding to rotations and reflections respectively
        \item The signal can also be obtained as a linear sum of all the low and high frequency components of each G-let
\end{itemize}

\subsection{\textit{Generate\_G-let\_Matrices()}}

\begin{itemize}
        \item \em$angle = 360^\circ /n$ taken in radians
        \item Calculate number and dimension of irreducible representations
        \item Identify the conjugacy class of matrices
        \item Make the transformation matrices R using, one and two dimensional irreducible representations placing them diagonally
        	    for rotations and reflections, in the case of Dihedral groups
        \item return R
\end{itemize}

\subsection{Reconstruction and Compression}
A linear sum of the G-let coefficients quickly allows for a perfect reconstruction of the discrete signal `sig' as in the equation \ref{Eq:linear_sum}.
\begin{equation}
sig = (G_1(sig) \oplus G_2(sig) \cdots \oplus G_{n-1}(sig))\times(-In).
\label{Eq:linear_sum}
\end{equation}

Reconstruction is also possible with just half the members of a conjugacy class that has $2$ members. For conjugacy classes of reflections, we cannot randomly choose a member since they are more than two. It is possible that among the members `m' of the reflection class, we can find the difference between the $m^{th}$ G-let and the sum of the other $m - 1$ G-lets in the class and use this difference to represent the specific conjugacy class of reflections. So the projections of G-lets one from each rotation class, along with the difference between two successive G-lets in the reflection class, will give us the original signal. We thus effectively need only one G-let from each conjugacy class. The original signal can be represented by simply, only so many G-let amplitudes or frequencies giving a lossless compression of approximately $ 50\%$ in terms of reconstructing the signal from all G-lets. The signal can also be reconstructed from just one G-let, since each G-let is a separate transform.

\section{G-lets for one-dimensional signals}
\begin{figure}[h]
\centering
\vfil
\subfigure[]{
\includegraphics[width=17pc]{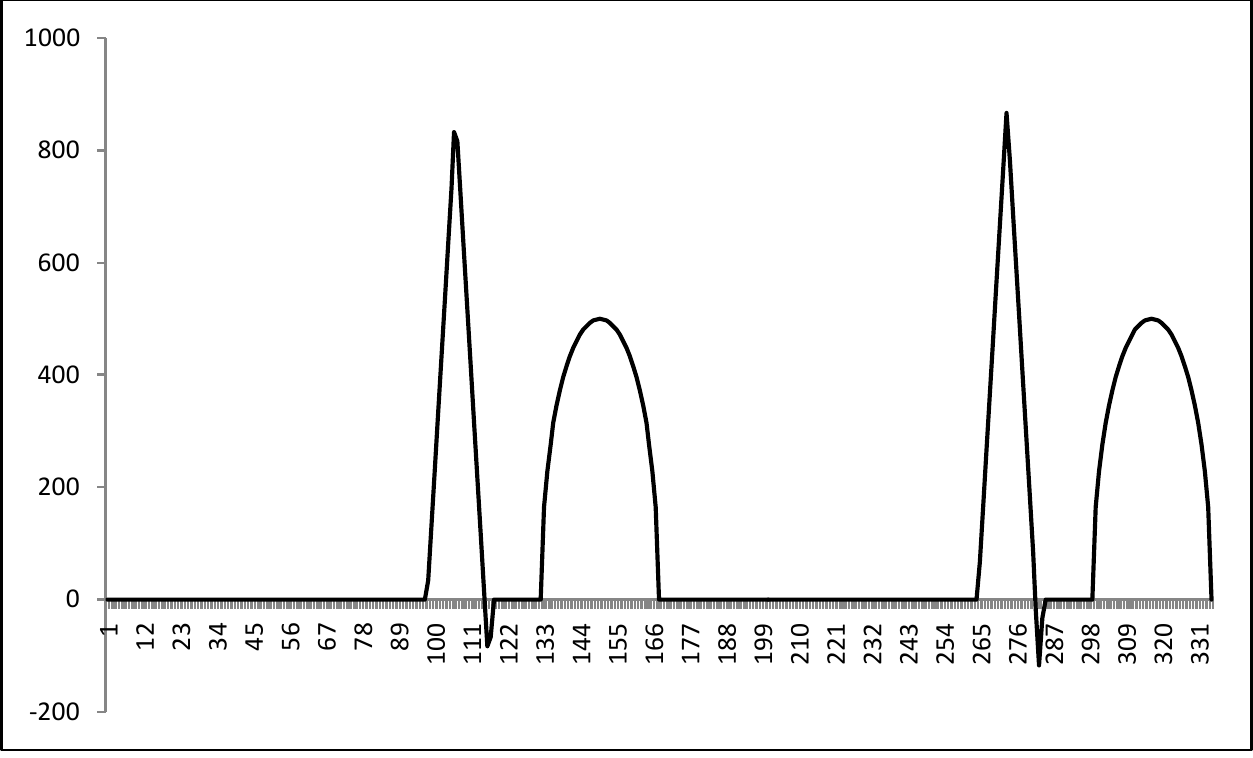}
\label{fig:original_ECG_signal}
}
\centering
\hfil
\subfigure[]{
\includegraphics[width=17pc]{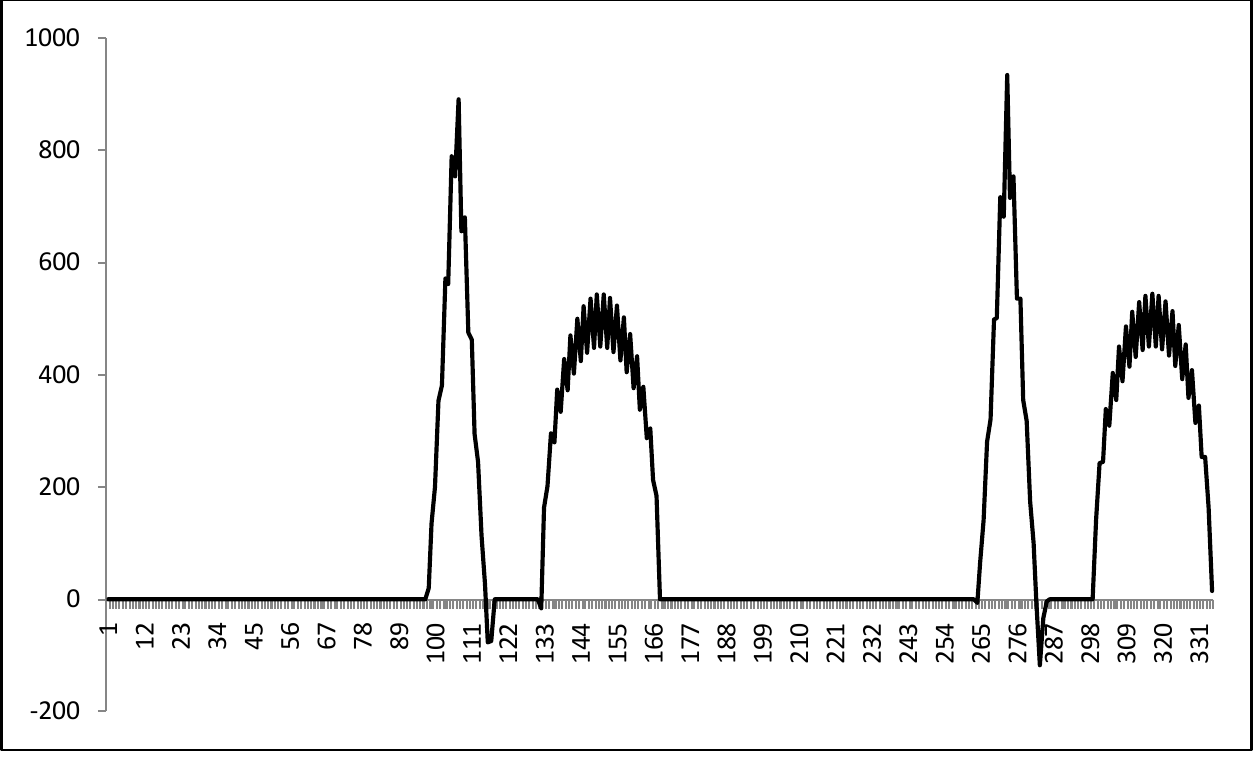}
\label{fig:5th G-let}
}
\centering
\vfil
\subfigure[]{
\includegraphics[width=17pc]{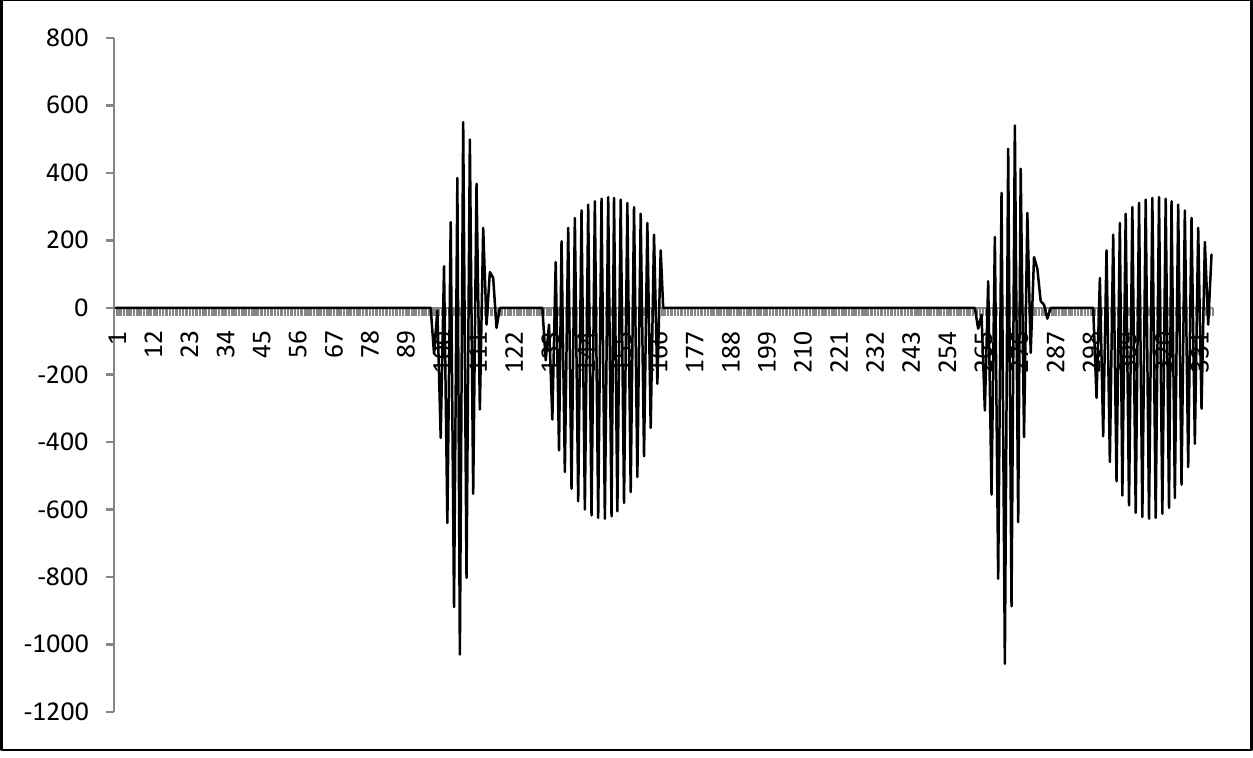}
\label{fig:100th_G-let}
}
\centering
\hfil
\subfigure[]{
\includegraphics[width=17pc]{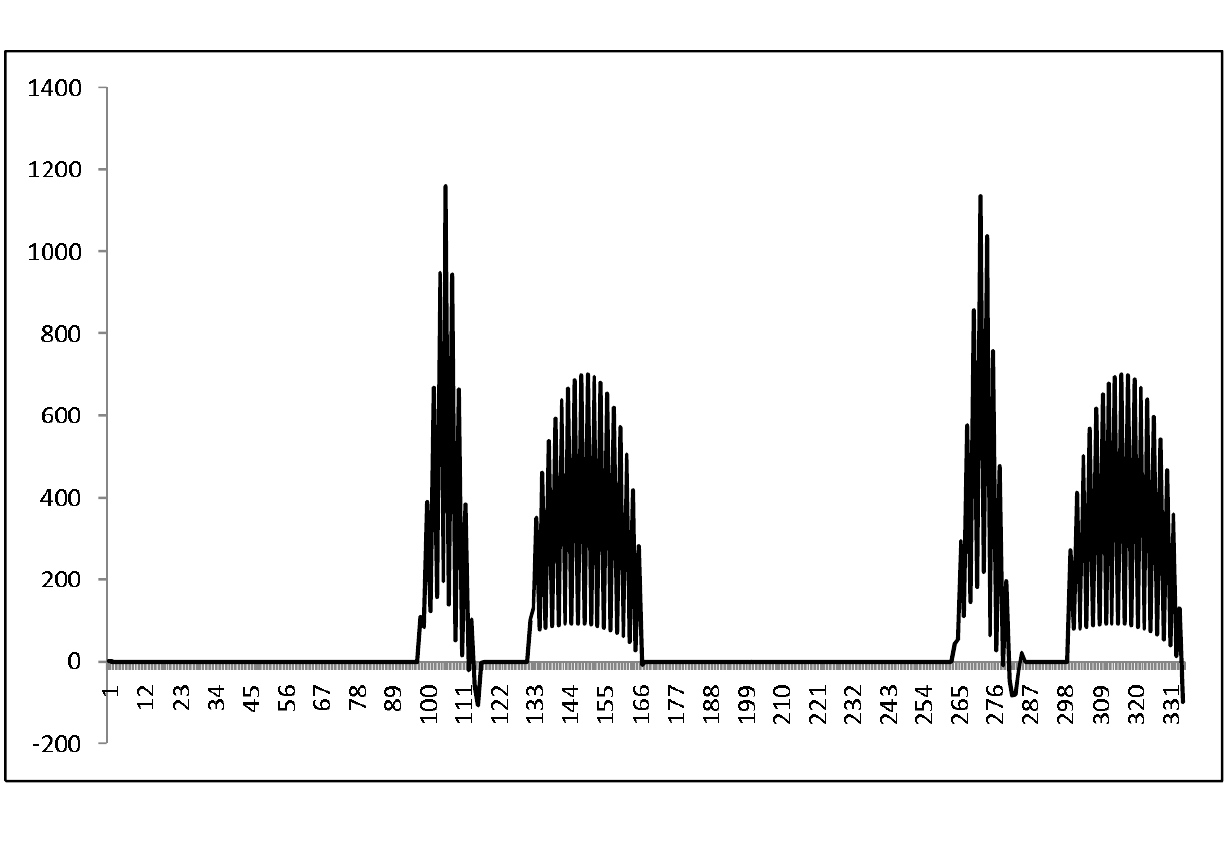}
\label{fig:300th_G-let}
}
\caption{(a)\ Original ECG signal\ (b)\ $5^{th}$ G-let\ (c)\ $100^{th}$ G-let}
\label{fig:ECG_results}
\end{figure}

Let us consider an ECG signal for demonstration. The chosen signal is of dimension $1 \times 336$. Therefore $n = 336$. The conjugacy classes are given by $(336+6)/2 = 171$. There are $169$ G-lets from all the classes. In other words, there are $167$ two dimensional irreducible representations and two one dimensional representations adding up to $n = 334+2$, the actual dimension of the signal. The signal is not projected separately for one dimensional representation as there is no significance. The ECG signal and some of its G-lets are shown in Fig.~\ref{fig:ECG_results}. The amplitudes and frequencies of each of these G-lets are also shown in Fig.~\ref{fig:ECG_results}. The $5^{th}$ and $100^{th}$ G-let of the ECG signal are shown in Fig.~\ref{fig:5th G-let} and Fig.~\ref{fig:100th_G-let}, respectively. The amplitude resolution of an ECG signal has been demonstrated in Fig.~\ref{fig:ECG_results}.

\section{G-lets for two-dimensional signal}

We now turn to analysis of images using G-lets. We consider the standard image `Lena'. The image has a dimension $n = 256$. The number of conjugacy classes is $(256+6)/2 = 131$. Four among these are identity representations. We get a count of $127$ G-lets from all the classes. This works out to 49.48$\%$ reduction for reconstruction of G-let images.

\subsection{G-let Analysis}

\begin{figure}[h]
\centering
\vfil
\subfigure[]{
\includegraphics[width=25pc]{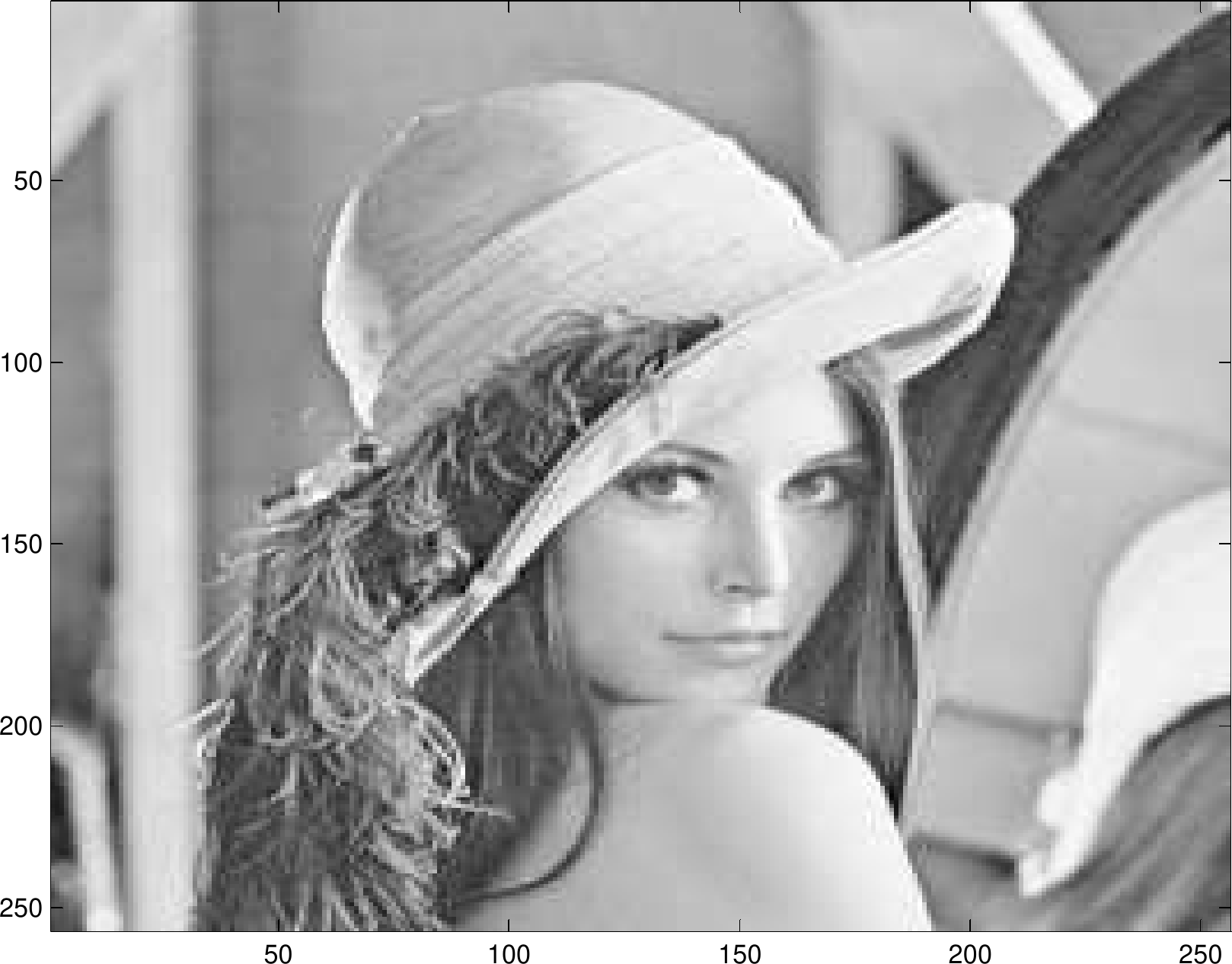}
\label{fig:lena}
}
\vfil
\subfigure[]{
\includegraphics[width=25pc]{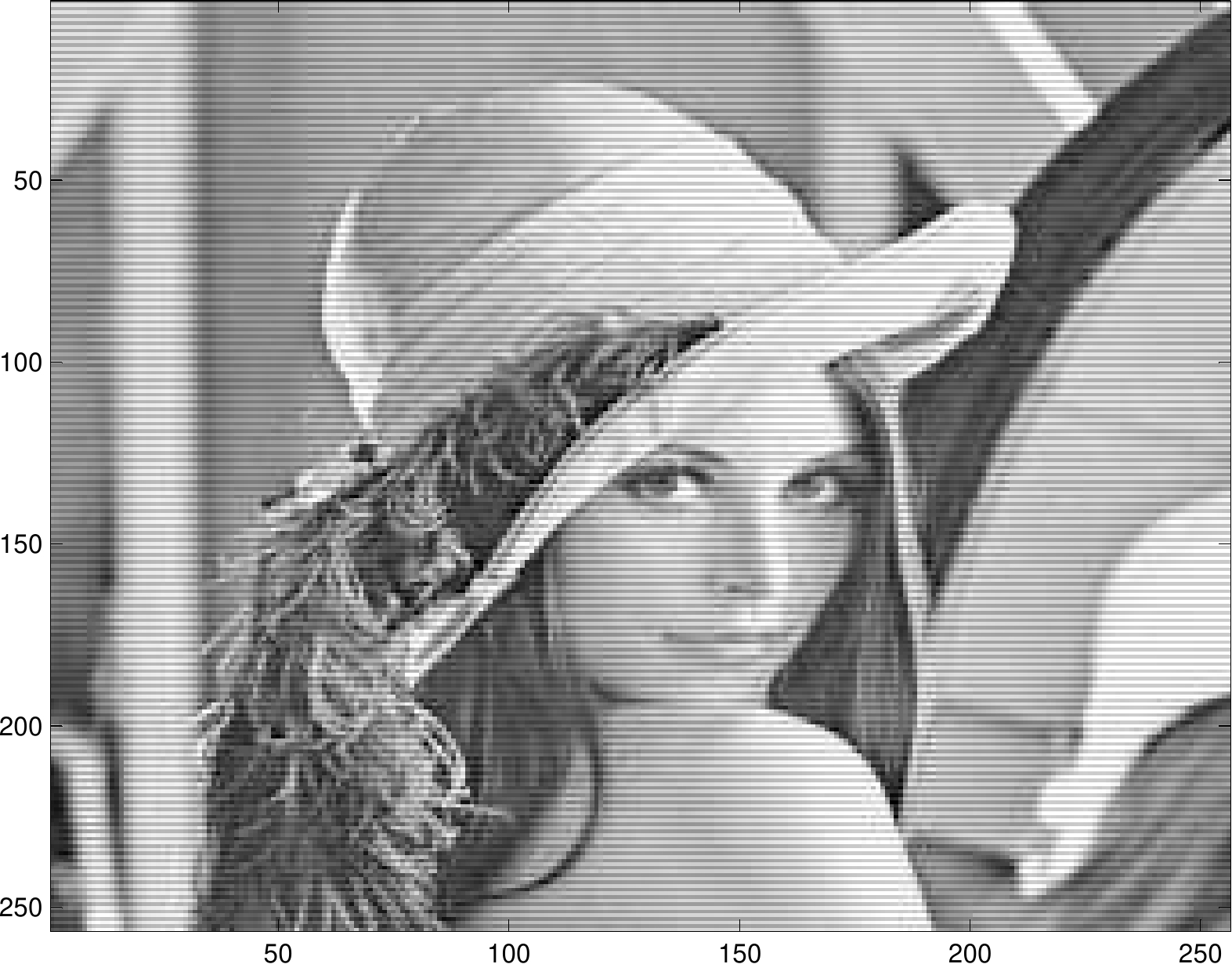}
\label{fig:lena10}
}
\caption{(a)\ `Lena' image\ ;\ \ Lena Rotation G-lets:\ \ (b)\ $10^{th}$ G-let}
\label{fig:lenarotation1}
\end{figure}

\begin{figure}[h]
\centering
\vfil
\subfigure[]{
\includegraphics[width=25pc]{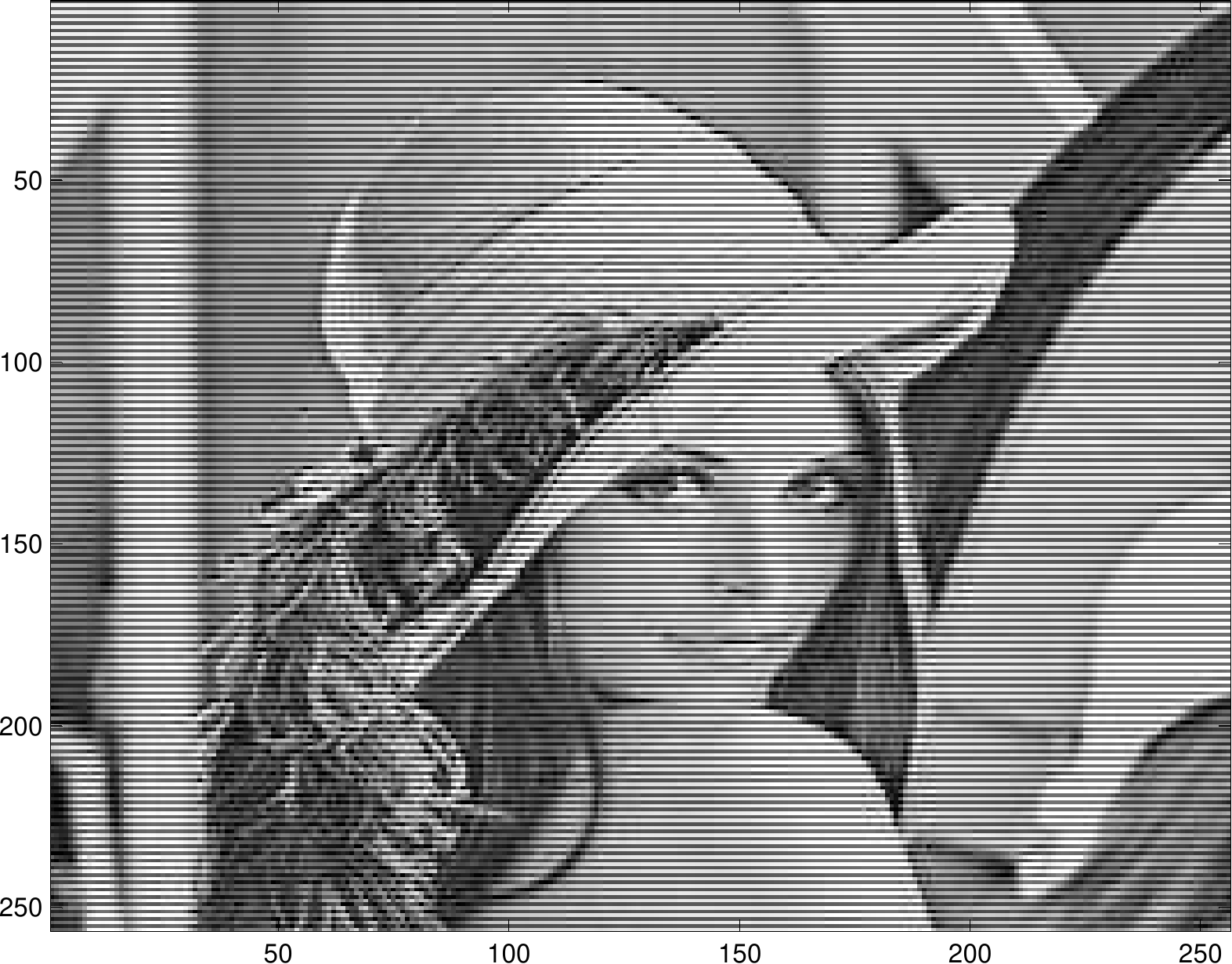}
\label{fig:lena25}
}
\vfil
\subfigure[]{
\includegraphics[width=25pc]{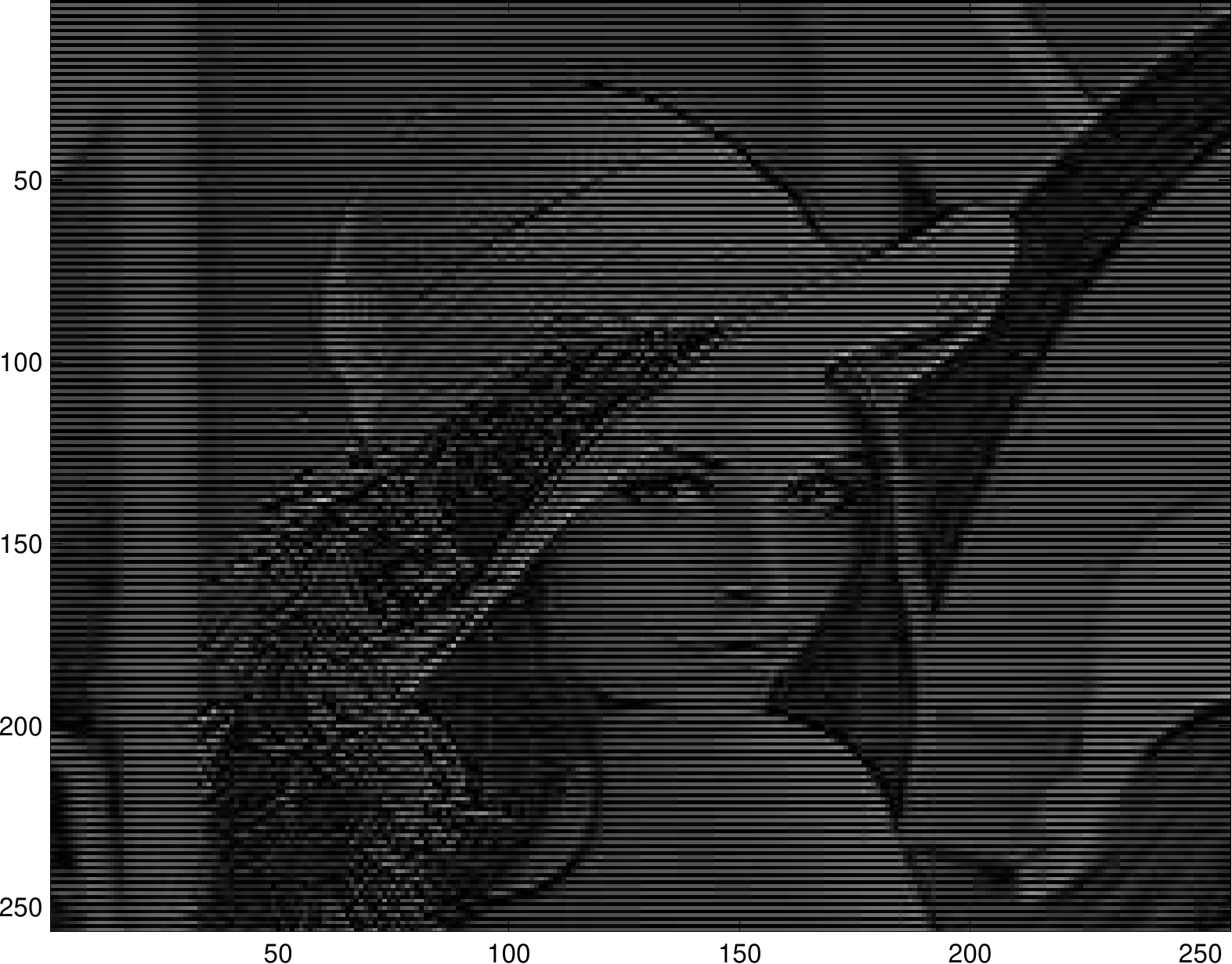}
\label{fig:lena89}
}
\caption{Lena Rotation G-lets:\ \ (a)\ $25^{th}$ G-let\ (b)\  $89^{th}$ G-let\ }
\label{fig:lenarotation2}
\end{figure}

\begin{figure}[h]
\centering
\hfil
\subfigure[]{
\includegraphics[width=25pc]{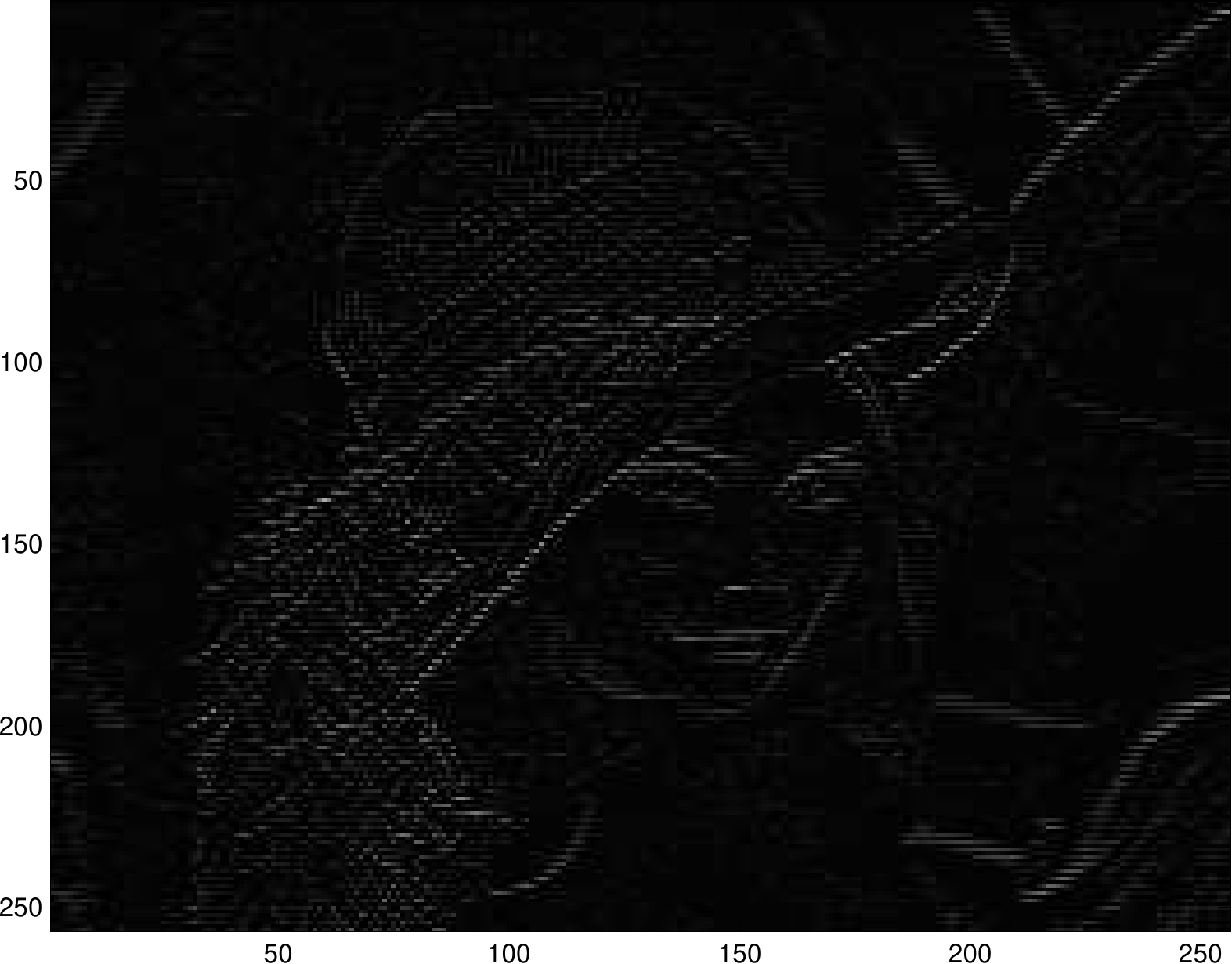}
\label{fig:lena96}
}
\vfil
\subfigure[]{
\includegraphics[width=25pc]{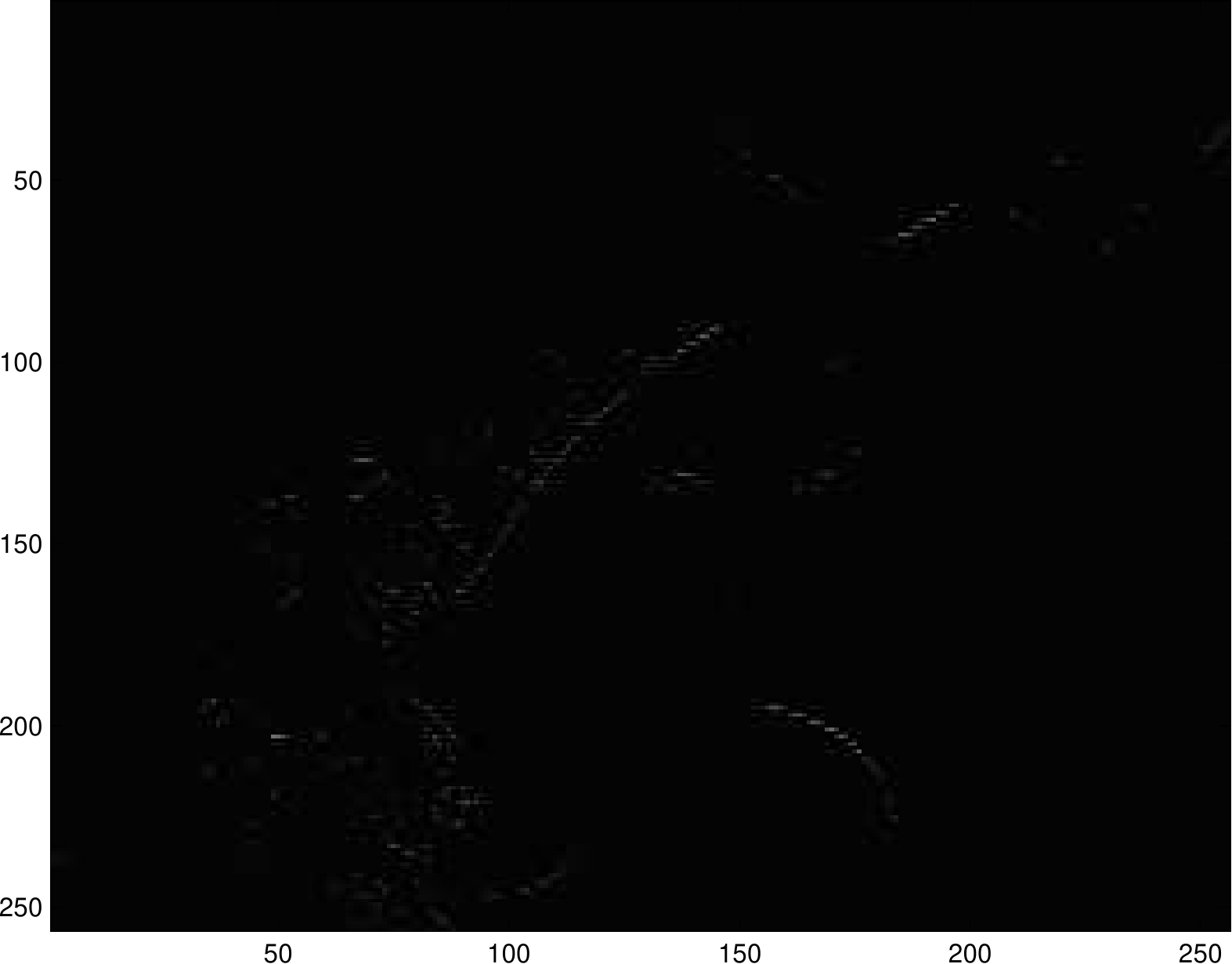}
\label{fig:lena150}
}
\caption{Lena Rotation G-lets:\ \ (a)\ $96^{th}$ G-let\ (b)\  $150^{th}$ G-let\ }
\label{fig:lenarotation3}
\end{figure}

\begin{figure}[h]
\centering
\hfil
\subfigure[]{
\includegraphics[width=25pc]{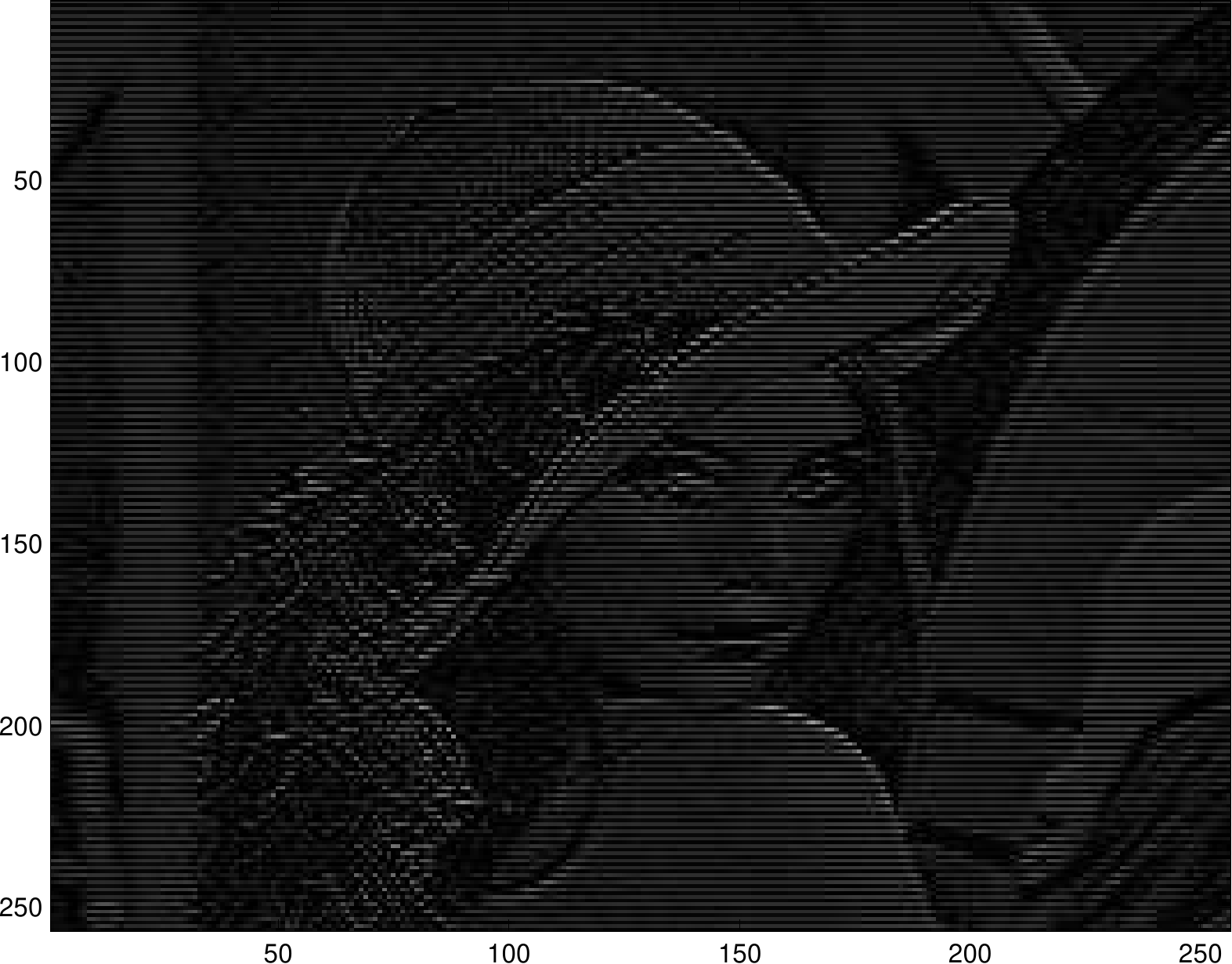}
\label{fig:lena162}
}
\caption{Lena Rotation G-lets:\ \ (a)\ $162^{nd}$ G-let}
\label{fig:lenarotation4}
\end{figure}

The `Lena' image and it amplitude resolution is shown in the figures Fig.~\ref{fig:lena},
Fig.~\ref{fig:lena10}, Fig.~\ref{fig:lena25}, Fig.~\ref{fig:lena89}, Fig.~\ref{fig:lena96}, Fig.~\ref{fig:lena150} and Fig.~\ref{fig:lena162}. The `Lena' image is of size $256$ x $256$.

Amplitudes: They are obtained from the G-let image by extracting the high positive amplitudes in each G-let. The amplitude coefficients of the image are shown in the Fig.~\ref{fig:lena10},Fig.~\ref{fig:lena25},Fig.~\ref{fig:lena89},Fig.~\ref{fig:lena96},Fig.~\ref{fig:lena150} and Fig.~\ref{fig:lena162}. The amplitudes are at different resolution in each G-let. In this way, a multiresolution analysis in terms of amplitudes is obtained by using the G-let matrices. In other words, only specific amplitudes of the signal are prominent in each G-let. Also with each G-let the resolution decreases. Finally the last G-let contains a totally reflected signal. An advantage of using transformation groups to produce amplitude resolution is that, there is a conjugate pair for each G-let. If one G-let works out amplitudes of odd points in the discrete signal, the corresponding conjugate G-let projects even points in the same signal. Thus amplitude resolution is obtained in two kinds (while using dihedral groups), one for both the alternate series of the signal. In this way, none of the signal features are lost in the process.

\begin{figure}[h]
\centering
\vfil
\subfigure[]{
\includegraphics[width=25pc]{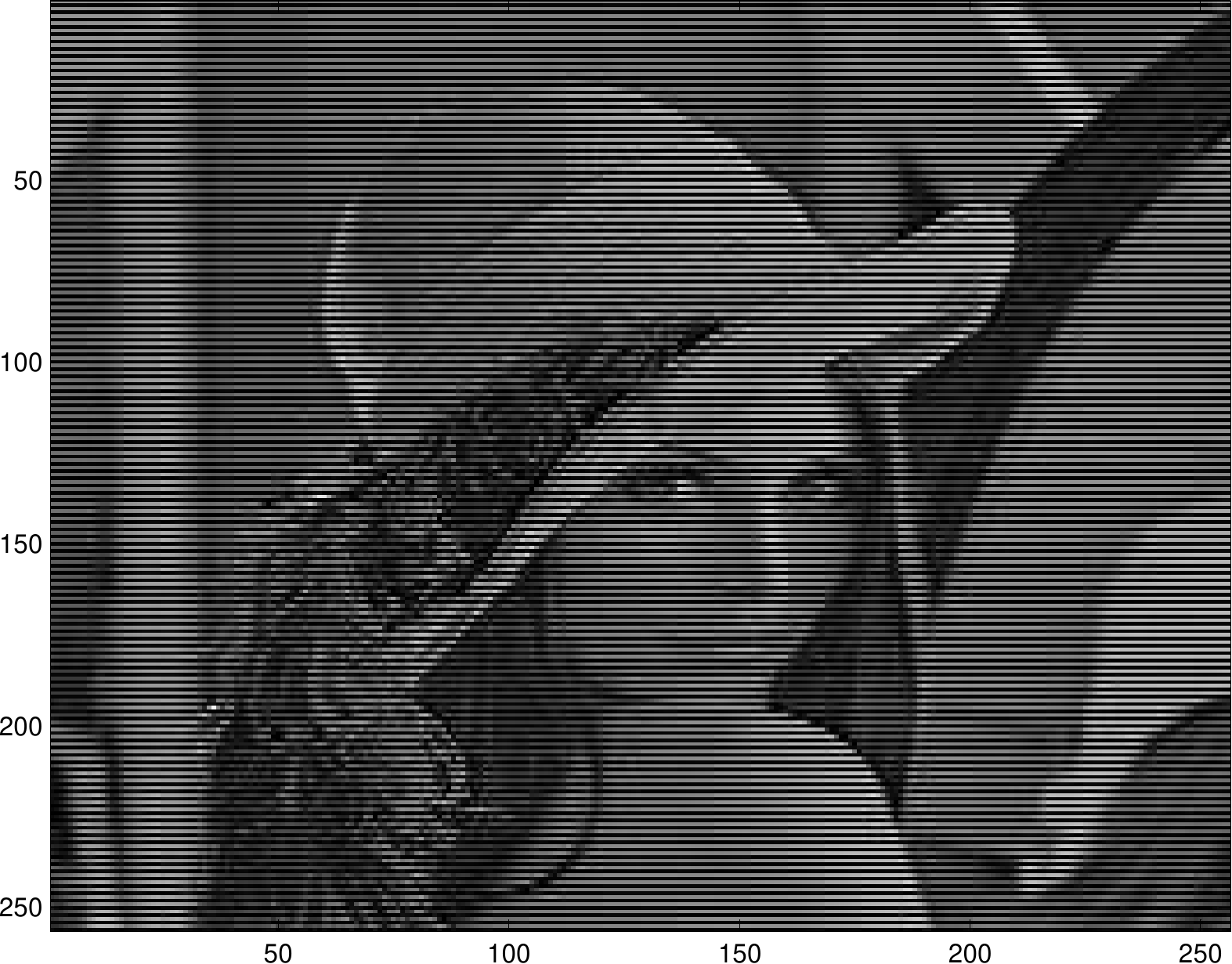}
\label{fig:lena14rf}
}
\vfil
\subfigure[]{
\includegraphics[width=25pc]{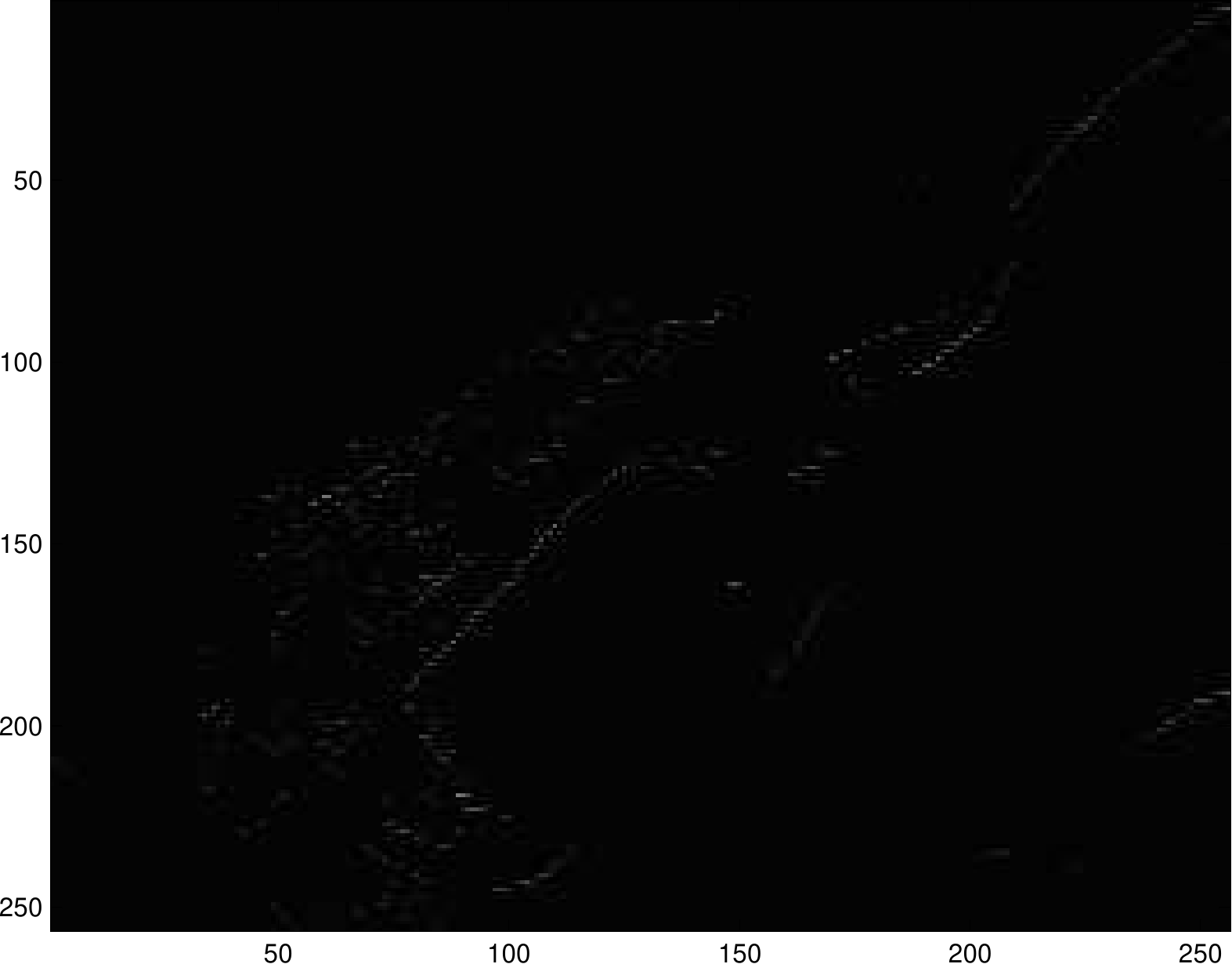}
\label{fig:lena41rf}
}
\caption{Lena Reflection G-lets:\ \ (a)\ $14^{th}$ G-let \ (b)\ $41^{st}$ G-let}
\label{fig:lenareflection1}
\end{figure}

\begin{figure}[h]
\centering
\vfil
\subfigure[]{
\includegraphics[width=25pc]{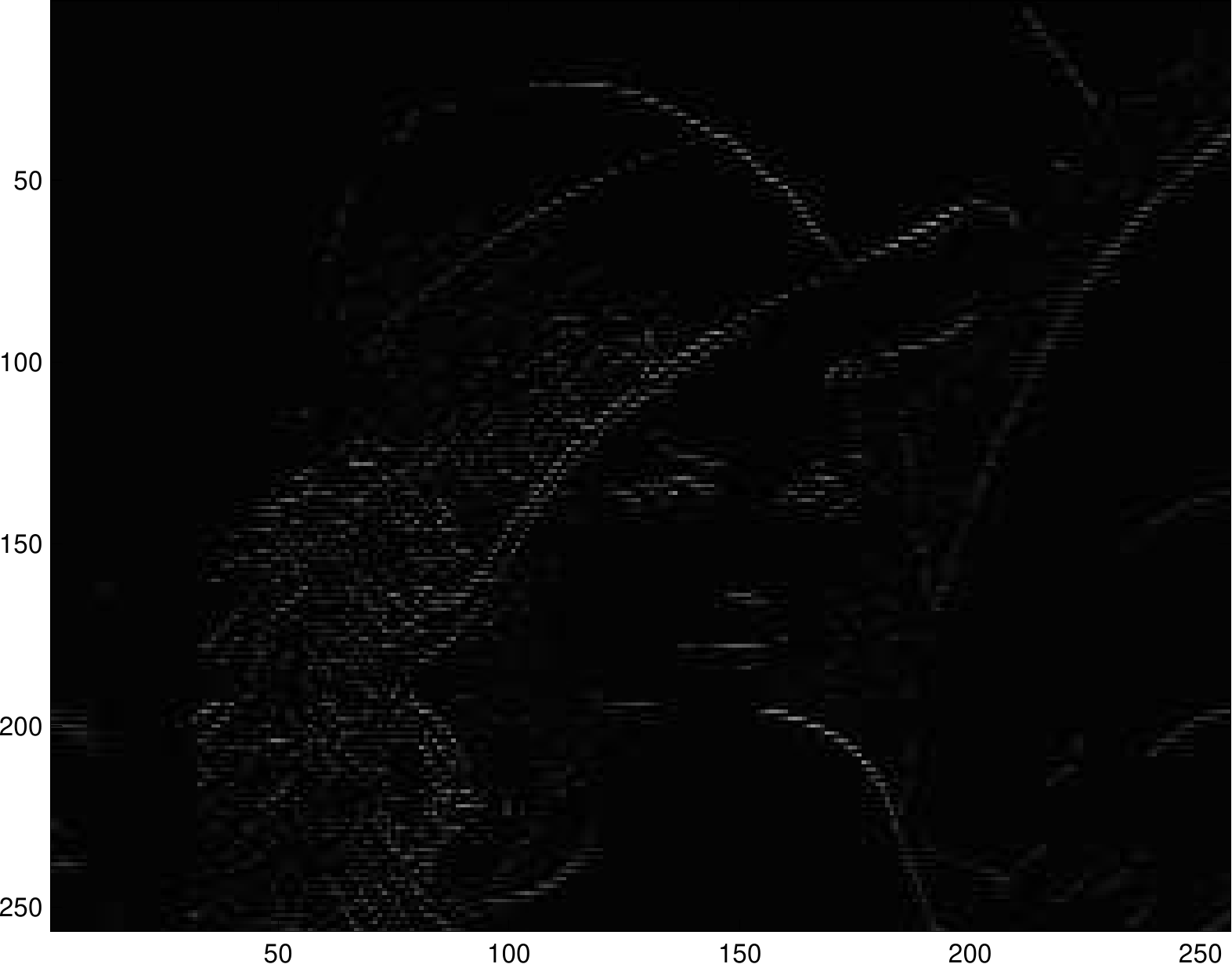}
\label{fig:lena84rf}
}
\vfil
\subfigure[]{
\includegraphics[width=25pc]{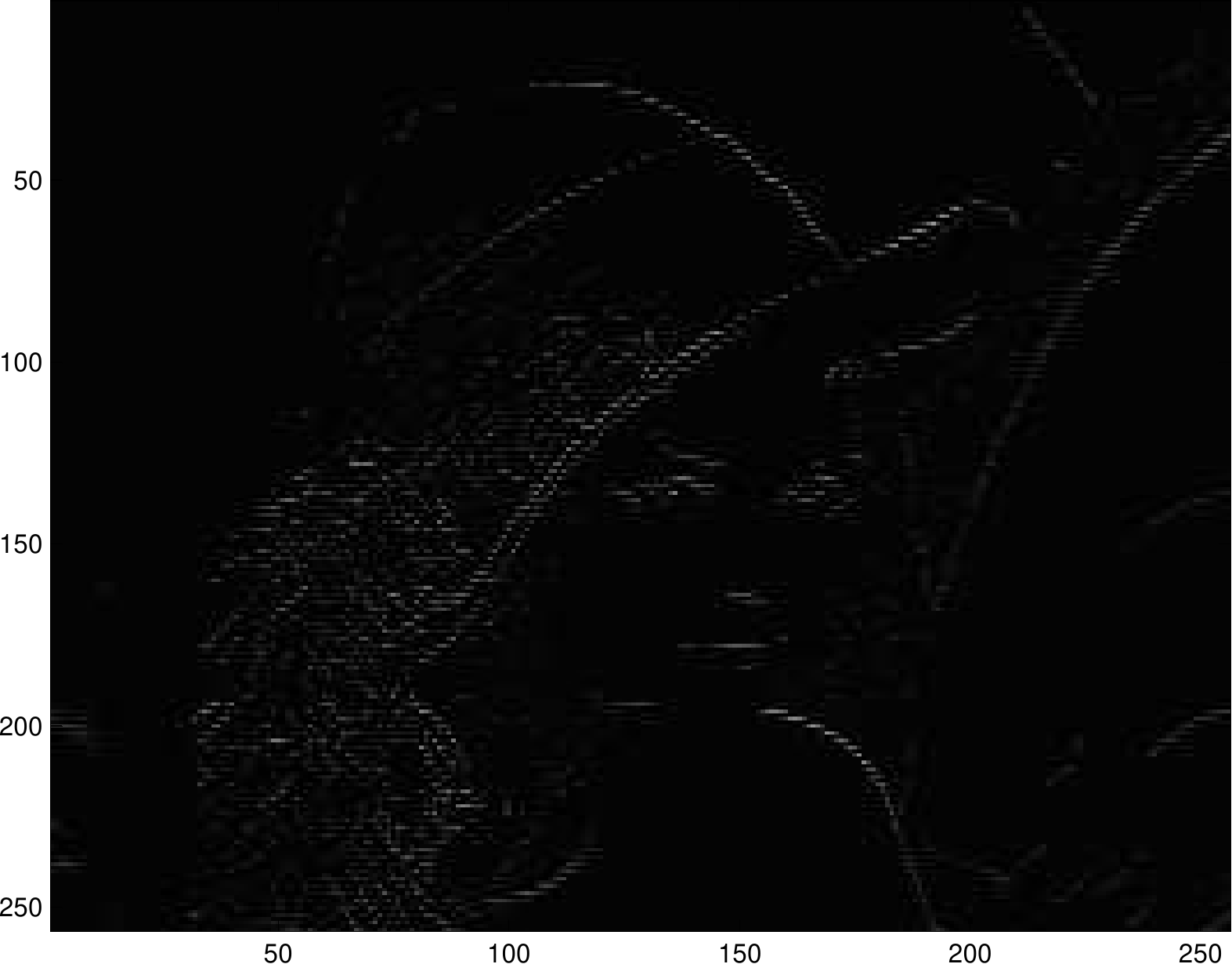}
\label{fig:lena94rf}
}
\caption{Lena Reflection G-lets:\ \ (a)\ $84^{th}$ G-let\ (b)\  $94^{th}$ G-let\ }
\label{fig:lenareflection2}
\end{figure}

\begin{figure}[h]
\centering
\hfil
\subfigure[]{
\includegraphics[width=25pc]{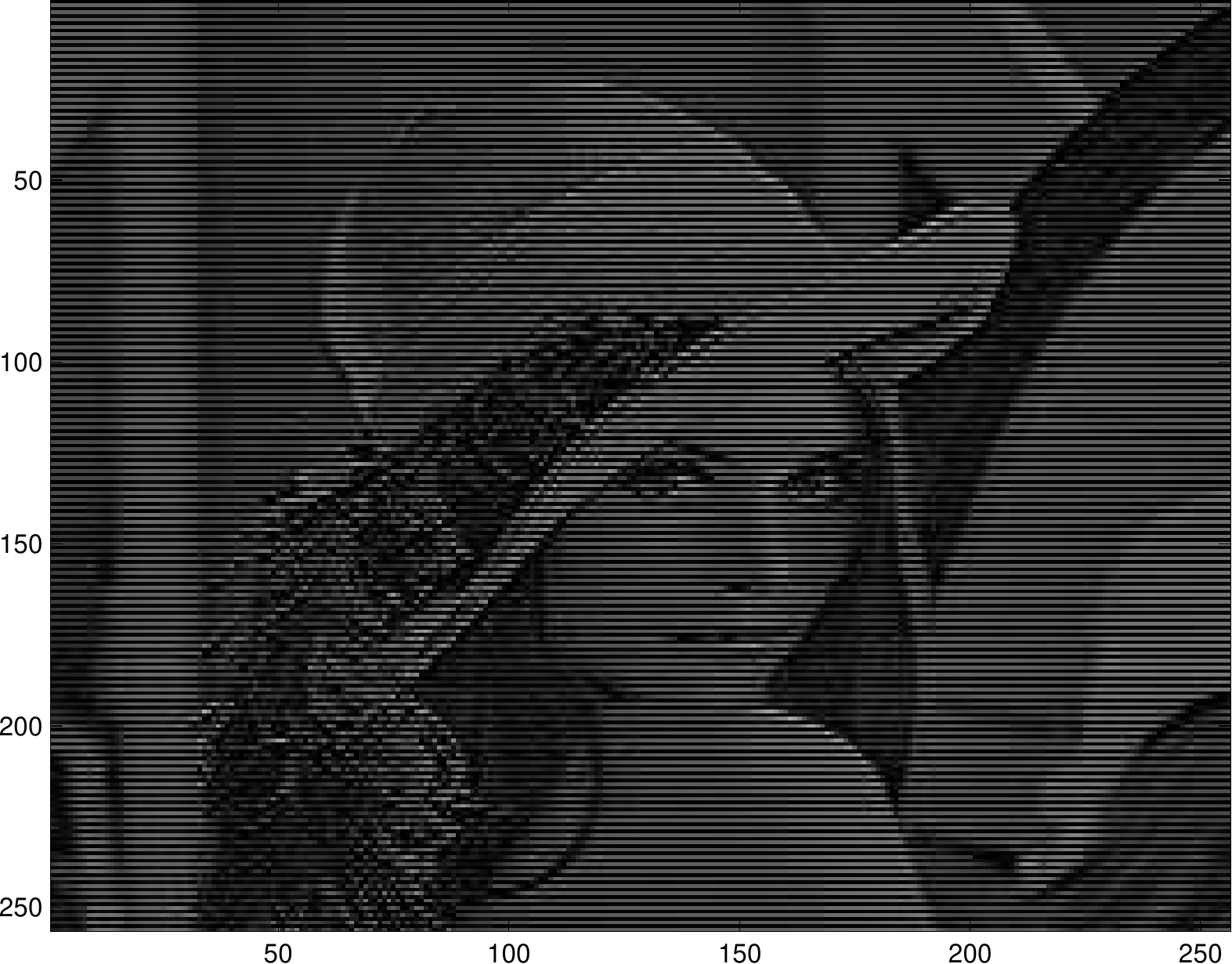}
\label{fig:lena104rf}
}
\vfil
\subfigure[]{
\includegraphics[width=25pc]{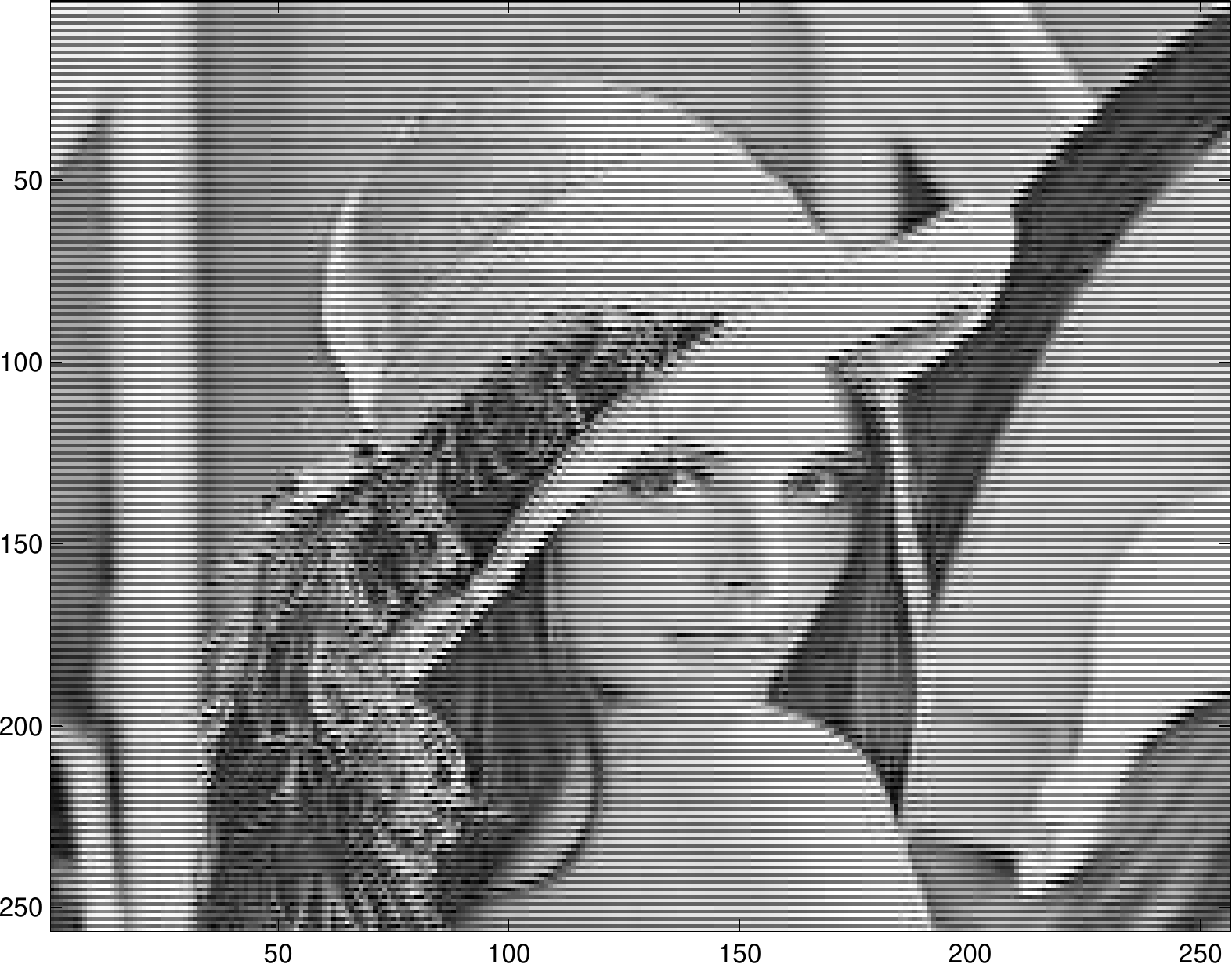}
\label{fig:lena169rf}
}
\caption{Lena Rotation G-lets:\ \ (a)\ $104^{th}$ G-let\ (b)\  $169^{th}$ G-let\ }
\label{fig:lenareflection3}
\end{figure}

Each G-let gradually decreases in its finer details during amplitude resolution. A linear sum of the signal projections on the G-lets gives back the original discrete signal with no loss. The corresponding amplitudes and frequencies show the local features at that degree of details of the signal. Let us say the dimension of the signal is n. There are 2n G-lets and they form two basis sets. Each set with their projections are used to reconstruct the image as depicted in the algorithm above. We see that the identity G-lets $G_n(sig)$ and $G_{2n}(sig)$ are also used in the reconstruction but not for the linear sum of G-let projections. Hence only $n - 1$ G-lets are actually used for signal projections in both the basis. The second set of G-lets (reflections), are $14^{th}$, $41^{st}$, $84^{th}$, $94^{th}$, $104^{th}$ and $169^{th}$ for the `Lena' image shown in figures Fig.~\ref{fig:lena14rf}, Fig.~\ref{fig:lena41rf}, Fig.~\ref{fig:lena84rf}, Fig.~\ref{fig:lena94rf}, Fig.~\ref{fig:lena104rf} and Fig.~\ref{fig:lena169rf}.

The lines seen in the image results are due to the two dimensional irreducible representations kept diagonally in the representation matrix.  As we can see the G-lets gradually reduce in details to edges and further down to points in the image which have very less depth in terms of $3D$ information. Before the image completely loses its points, it is found, the front layer of the image can be identified and as the details gradually appear again, we get the outline. The continuity in the points is a trend towards the growing outline of an object in the image whose limits are decided using a threshold. Depending on the threshold, the image is separated as layers from the front to the back. In each layer, the limits of the outline are boxed and all the G-lets are used to reconstruct that portion of the original image within the box. Every layer is put together to get the collective image back.

\section{Frequency Analysis}
The low and high frequencies of a signal or image are obtained from each G-let by handling the oscillations induced by G-let matrices. By removing the oscillations, low frequencies are obtained in the removed component. High frequencies remain in the signal after this process. The portions of a signal where oscillations are seen, in both the 1D signal and image, is removed for frequency analysis. The nature of the oscillations depends on the transformation group considered and the G-let matrices design. These oscillations separate out low frequencies and therefore high frequencies are filtered and settled in the signal. The smooth portions of the signal are perceived to be the low frequency components. Whereas the sharp regions of the signal are taken as high frequency components. For every G-let a low and high frequency split of the signal is produced. Some G-lets give the highest frequency of the signal with all other frequencies heavily suppressed. In the process of splitting the low and high frequency components, the time information of the frequencies is not lost. This gives rise to a simple time-frequency analysis method. As a result, a multiresolution analysis based on frequencies is obtained for the signal using G-lets.

Amplitude resolution of a signal happens at the first level using G-lets and at the second level, frequency resolution emerges. For each frequency resolution, amplitude resolution can be generated by reapplying G-let matrices. Elaborate discussion of time-frequency analysis by G-lets is being prepared as a separate manuscript.

\section{Comparison with Wavelet Analysis}

A mother wavelet is used for the generation of wavelet basis ~\cite{Ref:Mallatwavelet} ~\cite{Ref:Daubechies-1988} ~\cite{Ref:Daubechies} for a signal. The signal is first placed in the frame of a Quadtree, where the vertical details, horizontal details, diagonal details and an approximation of the signal constitute the frame. The wavelets are then generated at different levels with scales varying by $2^j$. Both the approximation and the details of the signal can be analyzed in further levels. The resolution of the signal varies at each level. In G-lets, due to its natural multiresolution character, there is no need for a data structure such as Quadtree to create multiresolution. One of the important consequences here is that there is also no need for an approximation through a signal like a mother wavelet. A hierarchy of resolutions can be made starting from any of the G-lets. We also obtain compression of the signal without resorting to an approximation. If the transformations used in wavelets are placed within a transformation group instead of using a Quadtree, we can get G-lets for that group even without the mother wavelet and still do multiresolution analysis.

Geometric wavelets~\cite{Ref:Chen} choose a plane to separate finer and coarser details. Using PCA(principal component analysis) on the finer details, multiscale analysis is obtained. Choosing such a plane is equivalent to choosing a specific transformation group in the context of G-lets. For different transformation groups, a different set of fine and coarser details are obtained with respect to amplitude resolution and frequency resolution. By reiterating G-let decomposition on one of these resolutions, multiscale analysis is performed instead of a PCA. Therefore G-lets is also a generalization of geometric wavelets.

The computational complexity of the G-let decomposition is also less. Since G-lets have sparse matrix representations of the transformation group, computational complexity depends mainly on the number of irreducible representations which is $(n+6)/2$ for even sized signal and $(n+3)/2$ for odd sized signal. Hence the complexity of G-let decomposition is $O(n)$.

\section{Comparison with Fourier Analysis}

Fourier transform uses the characters of the group representations. The regular Fourier transform is for Abelian groups where each element of the group forms a conjugacy class by itself because of their commutative nature. Finite Fourier analysis~\cite{Ref:Vale}~\cite{Ref:Viana}~\cite{Ref:Assefa}~\cite{Ref:Stankovic}  for non-abelian groups is also defined. Again the characters of the one and two dimensional representations are used for Fourier transform. In this work, the representations themselves are used directly instead of the characters. For each transformation in the chosen group, a representation matrix is generated. A signal is projected onto these representation matrices and the chosen projections summed up to reconstruct the signal. The set of projections chosen for reconstruction depends on the type of transformations considered in the group. In dihedral groups there are two types, hence two sets are available. Each projection also smoothly transits from one resolution to the other and there are complementary projections in each conjugacy class.

\section{Implications of G-lets}
We now state some of the implications of G-lets without resorting to detailed analysis of each of them.

\begin{itemize}

\item	
G-lets can be used as a signal processing method for many applications like edge detection, feature extraction, denoising, face recognition, etc.

\item	
The different views projected by the two basis sets of G-lets facilitate not only edge detection or face recognition of an image but a convenient way of fine tuning the extraction by looking at both the views.

\item	
Having more than one basis set for the signal and using this to look at the structural symmetries in terms of amplitudes and frequencies of a signal in different ways resembles the behavior of a kaleidoscope. Which means more basis sets should give us a better kaleidoscope. This may be achieved with custom transformation groups.

\item
Custom transformation groups may be generated with the rules of group theory and with different combinations of transformations. More than two transformations may also be combined, which will give us that many basis sets, and a better kaleidoscope.

\item
G-lets may be generated using other standard transformation groups. Such G-lets could be suitable to a specific application that matches the properties of the chosen group. For example, dilation groups, rotation\-translation groups, cyclic groups, etc. G-lets have been tested for simple dilation groups.

\item
The nature of conjugacy classes of the transformation group allows for a reconstruction of the signal with only the number of G-lets equal to the number of conjugacy classes. This trait allows for lossless compression of a discrete signal straight down to almost $50\%$ in terms of reconstructing the signal using G-lets.

\item	
Multiresolution analysis is the inherent nature of G-lets. It is possible using both amplitudes and frequencies separately. Multiscale analysis is also possible by repeatedly performing amplitude and frequency multiresolution analysis for each G-let hierarchically.

\item
If a new basis is generated for any one of the G-lets, only those portions of the signal that this G-let represents are expanded in the basis. In this way multiple levels of multiresolution analysis is possible for the signal. This is local multiresolution and again can be done in terms of amplitude or frequency.

\item
Customizing the arrangement of irreducible representation matrices in each G-let matrix, a variety of amplitude and frequency resolution is possible. For example, by using irreducible representations of different rotations in the same G-let matrix, the corresponding regions affected by them show different kinds of projecting amplitudes in each of them. In this paper, one rotation is included in one G-let matrix. That makes the effect uniform. But when this is varied, for applications like edge detection, diverse portions of a signal are evaluated in a different manner. Thereby minute edges are identified and become prominent in any one of the G-lets. Such G-let coefficients still make a linear sum and reconstruct the original signal.

\item
Since each G-let matrix is one transform, the signal can be reconstructed from just one G-let alone. By separating the low and high frequencies a low resolution version of the signal can be obtained, which shows compression of the signal from the chosen G-let.

\end{itemize}

\section{Conclusions}

We have explored how multiresolution analysis is possible for a signal using a basis (termed G-lets) generated from a group of transformations. Amplitudes and frequencies of images may be extracted easily from images, and the natural multiresolution analysis that arises from the basis allows for a number of applications including denoising, edge detection and face recognition. Separate manuscripts are being prepared for these applications which have been tested.

\end{document}